# A Comparison of Three Measurement Models for the Wheel-mounted MEMS IMU-based Dead Reckoning System

Yibin Wu, Xiaoji Niu and Jian Kuang

*Abstract*—A self-contained autonomous dead reckoning (DR) system is desired to complement the Global Navigation Satellite System (GNSS) for land vehicles, for which odometer-aided inertial navigation system (ODO/INS) is a classical solution. In this study, we use a wheel-mounted MEMS IMU (Wheel-IMU) to substitute the odometer, and further, investigate three types of measurement models, including the velocity measurement, displacement increment measurement, and contact point zero-velocity measurement, in the Wheel-IMU based DR system. The measurement produced by the Wheel-IMU along with the non-holonomic constraint (NHC) are fused with INS through an error-state extended Kalman filter (EKF). Theoretical discussion and field tests illustrate the feasibility and equivalence of the three measurements in terms of the overall DR performance. The maximum horizontal position drifts are all less than 2% of the total travelled distance. Additionally, the displacement increment measurement model is less sensitive to the lever arm error between the Wheel-IMU and the wheel center.

*Index Terms*—Wheel-mounted IMU, dead reckoning, odometer-aided INS, zero-velocity updates, vehicular navigation.

## NOMENCLATURE

a) Matrices are denoted in uppercase bold letters.
b) Vectors are denoted in lowercase bold italic letters.
c) Scalars are denoted in lowercase italic letters.
d) Coordinate frames involved in the vector transformation are denoted as superscript and subscript. For vectors, the superscript denotes the projected coordinate system.
e) $\hat{*}$ denotes the estimated or computed values.
f) $\tilde{*}$ denotes the observed or measured values.
g) $a_x$ denotes the element of vector $a$ on the $x$ axis.

## I. INTRODUCTION

THE Global Navigation Satellite System (GNSS) has been commonly used for vehicular navigation since its very birth. Although it can provide accurate positioning service in line of sight conditions [1], the stability and reliability deteriorate in complicated environments such as urban canyons and forests owing to the multipath and signal blockage [2, 3]. Therefore, other relative positioning systems are required to complement the GNSS to maintain the accuracy during GNSS outages.

When considering a self-contained autonomous navigation system which is immune to the disturbance from surroundings, it is not reasonable to rely on the exteroceptive sensors, e.g., camera and light detection and ranging (LiDAR) [4-8]. These visual navigation systems base on the perception of the external environments which suffer from the illumination variation, high motion blur, extreme weather conditions and etc.

The inertial navigation system (INS) is an old but widely used technology to determine the attitude and position for land vehicles. With the explosive development of microelectromechanical system (MEMS) techniques, MEMS inertial measurement units (IMUs) have been extensively applied for vehicular navigation owing to their low cost, small size, light weight and low power consumption [9]. Nonetheless, the positioning error of INS drifts quickly with time because of the significant inherent noise and bias instability, especially for low-end sensors. In consequence, other sensors are needed to limit the error accumulation of INS.

The odometer-aided INS (ODO/INS), using either velocity or travelled distance as measurement, has been exhaustively studied for decades [10]. Particularly, a land vehicle cannot move in the directions perpendicular to the forward direction in the vehicle frame in general [11]. This is known as the non-holonomic constraints (NHC). It was proven that odometer and NHC contribute significantly to restrain the error drift and enhance the INS stability [12, 13]. Zhao *et al* [14] proposed an adaptive two-stage Kalman filter to solve the problem that the changes of the odometer scale factor error and the process noise degrades the filtering performance. Wu *et al* [10, 15] analyzed the global observability for the self-calibration of ODO/INS integrated system and implemented the self-calibration procedure with the aid of GPS. Authors in [16] applied the state transformation extended Kalman filter in the INS/OD system and addressed the covariance-inconsistency problem. A comparison of loosely-coupled mode and tightly-coupled mode for ODO/INS was presented in [17], where the travelled distance is used as

This work was funded by the National Key Research and Development Program of China (No. 2016YFB0501800 and No. 2016YFB0502202).
The authors are with the GNSS Research Center, Wuhan University, Wuhan, China. {ybwu, xjniu, kuang}@whu.edu.cn.

observation. Authors in [18] used odometer distance measurement to integrate with INS in degraded GPS environments. An INS/laser Doppler velocimeter (LDV) integrated navigation algorithm was proposed in [19], in which the distance increment errors over a given time interval were treated as measurements to fully exploit the NHCs and LDV information per cycle. Ouyang *et al* [20] analyzed the error characteristics of the odometer pulses and investigated three measurement models in the ODO/INS integrated system, including pulse accumulation, pulse increment, and pulse velocity measurement. Field experiments showed that the standard pulse velocity measurement yields the best positioning accuracy.

However, the reliability of the odometer data depends on the road conditions and vehicle maneuvers. It is also challenging to fuse information from different systems because of hardware modification and data transfer synchronization problems [21].

In addition to installing external odometer or accessing the onboard wheel encoder of the vehicle, the wheel velocity can be obtained by mounting the IMU to the vehicle wheel. Let one axis of the IMU be parallel to the rotation axis, the wheel velocity can thereby be calculated using the gyroscope outputs of that axis and the wheel radius. Moreover, rotating the IMU around an axis with a constant speed can cancel the constant sensor bias errors to some extent, namely, the rotation modulation [9, 22, 23].

In our previous study [24], a wheel-mounted MEMS IMU (Wheel-IMU)-based dead reckoning (DR) system called Wheel-INS is proposed. In Wheel-INS, the IMU is placed on the non-steering wheel of the vehicle. Then the vehicle forward velocity computed by the gyroscope outputs and wheel radius is treated as an external observation with NHC to fuse with the strapdown INS. Experimental results have illustrated that the positioning and heading accuracy of Wheel-INS have been respectively improved by 23% and 15% against ODO/INS. Furthermore, Wheel-INS exhibits significant resilience to the gyroscope bias comparing with ODO/INS.

Besides the wheel velocity, the Wheel-IMU can produce the displacement increment measurement by integrating the vehicle velocity and attitude; this observation would be more accurate and stable than the traveled distance in the vehicle frame produced by the odometer, because the vehicle attitude is used to project the vehicle displacement to the navigation frame at every IMU data epoch (usually at 200 Hz), shown as Fig. 3. That is to say, the displacement measurement would be more reliable compared to the distance measurement used in ODO/INS, especially when the vehicle is turning.

In particular, the Wheel-IMU can be used to determine the contact point between the wheel and ground. Similar to the NHC, it is true that the velocity of the contact point on the wheel is zero if the vehicle does not slide on the ground or jump off the ground. Hence, the zero-velocity measurement can be employed [25]. The observation model is deduced in Section III-C. Zero velocity update (ZUPT) has been widely used as external observation to suppress the error drift of INS for pedestrian dead reckoning [26] and vehicular navigation [12, 27].

In conclusion, based on our previous study [24], this paper investigates and compares three kinds of measurement models in Wheel-INS.

1) Velocity measurement: the wheel velocity calculated by the wheel radius and the gyroscope readings of the Wheel-IMU.
2) Displacement increment measurement: the displacement increment of the vehicle in a certain period of time, calculated by integrating the vehicle velocity and attitude within the time frame.
3) Contact point zero-velocity measurement: the velocity of the contact point on the wheel with respect to the ground equaling to zero in general.

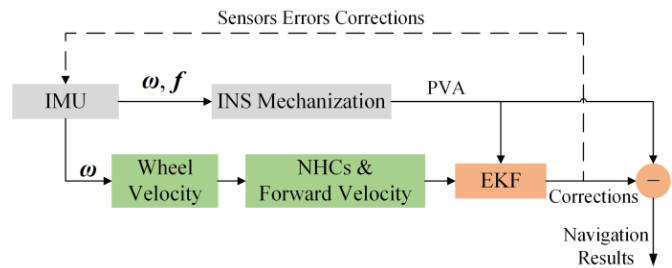

(a) System structure of the velocity measurement-based Wheel-INS [24].

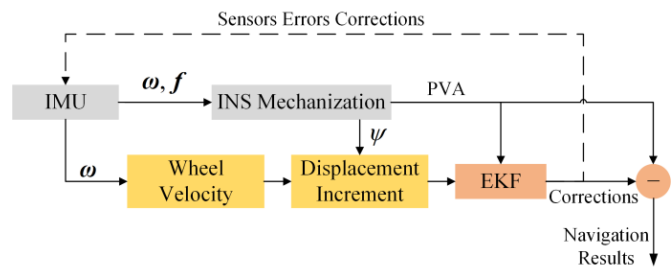

(b) System structure of the displacement increment measurement-based Wheel-INS.

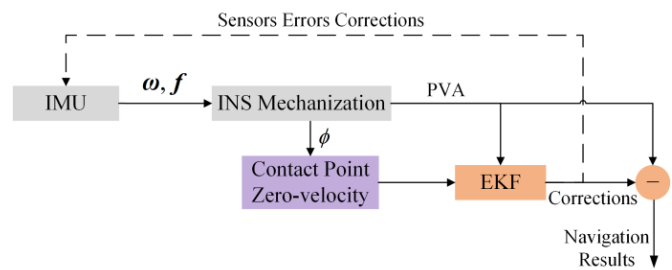

(c) System structure of the contact point zero-velocity measurement-based Wheel-INS.

Fig. 1 Overview of the structures of the three measurement models-based Wheel-INS. $\omega$ and $f$ are the angular rate and specific force measured by the Wheel-IMU, respectively; "PVA" indicates the position, velocity, and attitude of the IMU; $\psi$ indicates the vehicle heading; $\phi$ indicates the roll angle of the Wheel-IMU.

Fig. 1 depicts the algorithm flows of the three different measurement models-based Wheel-INS. The system is implemented using a 21 dimensional error-state extended Kalman filter (EKF). Details of the state model and observation models are presented in Section II–B and Section III, respectively. The state corrections estimated by the EKF are fed back to update the vehicle state and compensate the IMU outputs.

The remaining content is organized as follows. Section II gives the preliminaries of Wheel-INS, including the installation of the Wheel-IMU, the definition of the misalignment errors, and the error state model of the EKF. Section III deduces the three types of measurements and discusses their characteristic from the perspective of observation model. Experimental results are presented and analyzed in Section IV. Section V discusses the characteristics of the three measurement models. Section VI provides some conclusions and directions for future work.

## II. PREREQUISITES

Unlike the conventional ODO/INS system whereby the IMU is placed on the vehicle body, in Wheel-INS, the IMU is mounted on the wheel of the vehicle. In this section, the installation scheme of the Wheel-IMU and the coordinate systems are defined and analyzed firstly. Then we provide a review of the dynamic model of the error state adopted in the EKF to lay the foundation of Wheel-INS.

### A. Installation of the Wheel-IMU

To make the DR system indicate the vehicle state intuitively without being affected by vehicle maneuvers, the IMU is placed on a non-steering wheel of the vehicle. Fig. 2 illustrates the installation of the Wheel-IMU and the definition of the coordinate systems.

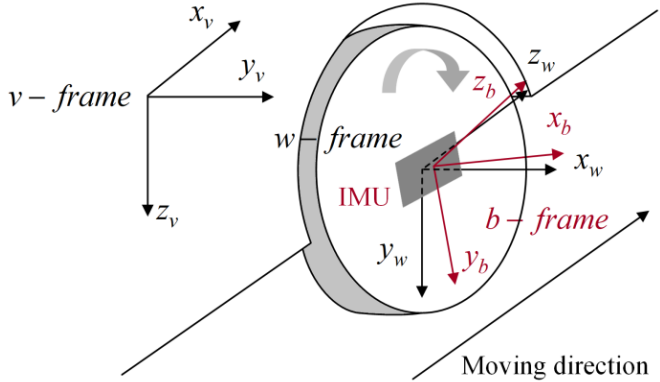

Fig. 2 Definition of the axes directions for the vehicle frame (*v*-frame), wheel frame (*w*-frame), and IMU body frame (*b*-frame). The position and attitude misalignment errors between the *b*-frame and the *w*-frame are also depicted [24].

The *v*-frame denotes the vehicle coordinate system, with the *x*-axis pointing to the advancement direction of the host vehicle, *y*-axis pointing right, *z*-axis pointing down, i.e., forward-right-down system. The *w*-frame denotes the wheel coordinate system. Its origin is at the rotation center of the wheel. Its *x*-axis points to the right of the vehicle, and its *y*- and *z*-axes are parallel to the wheel surface to complete a right-handed orthogonal frame. The *b*-frame denotes the IMU coordinate system, in which the accelerations and angular rates generated by the strapdown accelerometers and gyroscopes are resolved [28]. The *b*-frame axes are the same as the IMU's body axes. The *x*-axis is aligned with the wheel rotation axis, pointing to the right of the vehicle. Therefore, given a stable axle structure, the heading difference between the Wheel-IMU and the vehicle can be approximated as fixed (equaling to 90°), i.e.,

$$\psi_b^n = \psi_v^n + \pi/2 \tag{1}$$

where $\psi_b^n$ and $\psi_v^n$ denote the IMU heading and vehicle heading, respectively. *n* indicates the *n*-frame, which is a local-level frame with origin coinciding with the *b*-frame, *x*-axis directs at the geodetic north, *y*-axis east, and *z*-axis downward vertically, namely, the north-east-down system.

As shown in Fig. 2, it is inevitable that the *b*-frame is misaligned with the *w*-frame. Both the position (i.e., lever arm) and attitude misalignment errors (i.e., mounting angles) have been defined and analyzed in our previous study [24]. It is emphasized that these errors must be compensated in advance to obtain more robust and accurate state estimates. The lever arm can be measured or estimated online by augmenting it into the state vector, whereas the mounting angles can be calibrated by purpose-designed algorithm. Refer to [29] for details of the calibration procedure and error analysis.

The transformation from the *b*-frame to the *n*-frame can be written as

$$\mathbf{C}_b^n = \begin{bmatrix} c\theta c\psi & -c\phi s\psi + s\phi s\theta c\psi & s\phi s\psi + c\phi s\theta c\psi \\ c\theta s\psi & c\phi c\psi + s\phi s\theta s\psi & -s\phi c\psi + c\phi s\theta s\psi \\ -s\theta & s\phi c\theta & c\phi c\theta \end{bmatrix} \tag{2}$$

where $\mathbf{C}_b^n$ indicates the direction cosine matrix (DCM) from the *b*-frame to the *n*-frame; c and s indicate "cos" and "sin", respectively; $\phi$, $\theta$, and $\psi$ indicate the roll, pitch, and heading angle of the IMU, respectively. The transformation from the *b*-frame to the *w*-frame can be written as

$$\mathbf{C}_b^w = \begin{bmatrix} \cos\delta\theta\cos\delta\psi & -\sin\delta\psi & \sin\delta\theta\cos\delta\psi \\ \cos\delta\theta\sin\delta\psi & \cos\delta\psi & \sin\delta\theta\sin\delta\psi \\ -\sin\delta\theta & 0 & \cos\delta\theta \end{bmatrix} \tag{3}$$

where $\mathbf{C}_b^w$ indicates the DCM from the *b*-frame to the *w*-frame; $\delta\theta$ and $\delta\psi$ indicate the pitch and heading mounting angles of the Wheel-IMU with respect to the wheel frame, respectively. As the Wheel-IMU rotates with the wheel, the roll heading angle is not considered. Assume that the vehicle is moving on a horizontal plane and the mounting angles of the Wheel-IMU have been compensated, the transformation from the *n*-frame to the *v*-frame can be written as

$$\mathbf{C}_n^v = \begin{bmatrix} \cos\psi_v^n & -\sin\psi_v^n & 0 \\ \sin\psi_v^n & \cos\psi_v^n & 0 \\ 0 & 0 & 1 \end{bmatrix} \tag{4}$$

where $\mathbf{C}_n^v$ indicates the DCM from the *n*-frame to the *v*-frame; $\psi_v^n$ indicates the heading angle of the vehicle. As the vehicle is assumed to move on the horizontal surface, its pitch and roll angles are zero. With the transformation matrices presented above, all the transformation matrices between these four coordinates systems can be consequently determined.

Assuming the misalignment error has been calibrated and compensated, with the rotation of the wheel, the constant error of the gyroscope measurements in the two axes parallel to the wheel plane would be modulated into sine waves. After a period of integration, the accumulated heading error caused by the constant gyroscope bias error is canceled. [24] gives a heuristic explanation of the rotation modulation effect. Details can be found in [9, 30].

### B. Error State Model

In this study, the conventional strapdown INS is leveraged to predict the IMU state. The kinematic equations of INS are described at length in the literature [12, 23, 31]; thus, we do not go into details here. Additionally, we adopt the 21 dimensional error-state EKF to fuse the measurements with INS in Wheel-INS.

A large amount of state estimators for nonlinear systems have been proposed and applied to real world applications. Wheel-INS is a local DR system without the awareness of absolute heading and localization. Benefit from the rotation modulation, the heading drift of Wheel-INS is rather slow. Additionally, it can be observed from Eq. (6)-(8) that the state model in Wheel-INS is linear and quite simple. By deriving the error-state dynamics via perturbation of the nonlinear plant, the error-state EKF lends itself to optimal estimation of the error states [32, 33]. Therefore, those sophisticated state estimators like unscented Kalman filter (UKF) [34], particle filter (PF) [35], and strong tracking Kalman filter (STKF) [36, 37] would only limitedly improve the performance but increase computational cost instead. For the sake of simplicity and efficiency, we use the error-state EKF to implement the information fusion and state estimation in Wheel-INS. Moreover, we have proved in our latest paper [24] that the 21-state exhibits a better performance in Wheel-INS.

In this study, the state vector is constructed in the $n$-frame, including three dimensional position errors, three dimensional velocity errors, attitude errors, residual bias and scale factor errors of the gyroscope and accelerometer. It can be written as

$$\boldsymbol{x}(t) = \left[ \left( \delta \boldsymbol{r}^n \right)^T \quad \left( \delta \boldsymbol{v}^n \right)^T \quad \boldsymbol{\phi}^T \quad \delta \boldsymbol{b}_g^T \quad \delta \boldsymbol{b}_a^T \quad \delta \boldsymbol{s}_g^T \quad \delta \boldsymbol{s}_a^T \right]^T \quad (5)$$

where $\delta \boldsymbol{r}^n$, $\delta \boldsymbol{v}^n$, and $\boldsymbol{\phi}$ are the INS indicated position, velocity, and attitude errors, respectively; $\delta \boldsymbol{b}_g$ and $\delta \boldsymbol{b}_a$ are the residual bias errors of the gyroscope and the accelerometer, respectively; $\delta \boldsymbol{s}_g$ and $\delta \boldsymbol{s}_a$ are the residual scale factor errors of the gyroscope and accelerometer, respectively. Because of the errors from the sensors, IMU initial state, and other sources, the navigation parameters calculated by the INS mechanization equations contain errors. Several models have been developed to describe the time-dependent behavior of these errors [12]; the Phi-angle model is applied here, which can be expressed as

$$\dot{\boldsymbol{\phi}} = -\mathbf{C}_b^n \delta \boldsymbol{\omega}_{ib}^b \quad (6)$$

$$\delta \dot{\boldsymbol{v}}^n = \mathbf{C}_b^n \delta \boldsymbol{f}^b + \mathbf{C}_b^n \boldsymbol{f}^b \times \boldsymbol{\phi} \quad (7)$$

$$\delta \dot{\boldsymbol{r}}^n = \delta \boldsymbol{v}^n \quad (8)$$

where $\delta \boldsymbol{\omega}_{ib}^b$ and $\delta \boldsymbol{f}^b$ are the error vectors of the gyroscope and accelerometer, respectively, which can be expressed as $\delta \boldsymbol{\omega}_{ib}^b = \boldsymbol{b}_g + diag(\boldsymbol{\omega}_{ib}^b)\boldsymbol{s}_g$ and $\delta \boldsymbol{f}^b = \boldsymbol{b}_a + diag(\boldsymbol{f}^b)\boldsymbol{s}_a$; $diag(\bullet)$ is the diagonal matrix form of a vector; $\delta \boldsymbol{g}^n$ is the local gravity error in the $n$-frame. The sensor errors must be modeled to be augmented into the state vector. In this study, we chose the first-order Gauss-Markov process [38, 39] to model the residual sensor errors. The continuous-time model and discrete-time model are written as

$$\dot{x} = -\frac{1}{T}x + w$$
$$x_{k+1} = e^{-\Delta t_{k+1}/T} x_k + w_k \quad (9)$$

where $x$ is the random variable; $T$ is the correlation time of the process; $k$ is the discrete time index; and $w$ is the driving white noise. The continuous-time dynamic model and Jacobian matrix of the EKF can be found in [24].

## III. MEASUREMENT MODELS

In this section, the three different types of measurement models based on the Wheel-IMU are deduced. As discussed in Section-II-A, the misalignment errors of the Wheel-IMU can cause significant observation errors; they have to be calibrated previously for better performance. Here we assume that the lever arm is measured and the mounting angles are compensated in advance. Firstly, the vehicle forward velocity measurement produced by the gyroscope outputs and wheel radius is derived. Then, the displacement increment measurement using the vehicle heading to project the traveled distance in the $v$-frame to the $n$-frame is developed. Lastly, details of the construction of the contact point zero-velocity measurement is presented.

### A. Velocity Measurement

The wheel velocity indicated by the Wheel-IMU can be written as

$$\tilde{v}_{wheel}^v = \tilde{\omega}_x r - e_v = (\omega_x + \delta \omega_x) r - e_v$$
$$= v_{wheel}^v + r \delta \omega_x - e_v \quad (10)$$

where $\tilde{v}_{wheel}^v$ and $v_{wheel}^v$ indicate the observed and true wheel speed, respectively; $\tilde{\omega}_x$ is the gyroscope output in the x-axis; $\omega_x$ is the true value of the angular rate in the $x$-axis of the IMU; $\delta \omega_x$ is the gyroscope measurement error; $r$ is the wheel radius, and $e_v$ is the observation noise, modeled as the white Gaussian noise.

The motion of the wheeled robots is generally governed by two non-holonomic constraints [10, 40], which refers to the fact that the velocity of the robot in the plane perpendicular to the forward direction in the $v$-frame is almost zero [11, 12]. By integrating with the NHC, the 3 dimensional velocity observation in the $v$-frame can be expressed as

$$\tilde{\boldsymbol{v}}_{wheel}^v = \begin{bmatrix} \tilde{v}_{wheel}^v & 0 & 0 \end{bmatrix}^T - \boldsymbol{e}_v \quad (11)$$

Because the Wheel-IMU rotates with the wheel, the roll angle with respect to the wheel changes periodically. That is to say, it cannot be determined whether the vehicle is moving uphill or downhill by the Wheel-IMU alone. Therefore, we have to assume that the vehicle is moving on the horizontal plane. Nonetheless, experimental results in [24] have shown that it would not cause significant error if there are some degrees of slope in the road. According to Eq. (1), the Euler angles of the vehicle can be represented as

$$\boldsymbol{\varphi}_v^n = \begin{bmatrix} \phi_v^n \\ \theta_v^n \\ \psi_v^n \end{bmatrix} = \begin{bmatrix} 0 \\ 0 \\ \psi_b^n - \pi/2 \end{bmatrix} \quad (12)$$

where $\phi$, $\theta$, and $\psi$ are the roll, pitch, and heading angle of the vehicle, respectively.

By performing the perturbation analysis, the INS-indicated velocity in the $v$-frame can be written as

$$\begin{aligned}
\hat{\boldsymbol{v}}_{wheel}^v &= \hat{\mathbf{C}}_n^v \hat{\boldsymbol{v}}_{IMU}^n + \hat{\mathbf{C}}_n^v \hat{\mathbf{C}}_b^n \left( \hat{\boldsymbol{\omega}}_{nb}^b \times \right) \boldsymbol{l}_{wheel}^b \\
&\approx \mathbf{C}_n^v \left( \mathbf{I} + \delta \boldsymbol{\psi} \times \right) \left( \boldsymbol{v}_{IMU}^n + \delta \boldsymbol{v}^n \right) \\
&\quad + \mathbf{C}_n^v \left( \mathbf{I} + \delta \boldsymbol{\psi} \times \right)(\mathbf{I} - \boldsymbol{\phi} \times) \mathbf{C}_b^n \left( \boldsymbol{\omega}_{nb}^b \times + \delta \boldsymbol{\omega}_{ib}^b \times \right) \boldsymbol{l}_{wheel}^b \\
&\approx \boldsymbol{v}_{wheel}^v + \mathbf{C}_n^v \delta \boldsymbol{v}^n + \mathbf{C}_n^v \left[ \left( \mathbf{C}_b^n \left( \boldsymbol{\omega}_{nb}^b \times \right) \boldsymbol{l}_{wheel}^b \right) \times \right] \boldsymbol{\phi} \\
&\quad - \mathbf{C}_n^v \left[ \left( \boldsymbol{v}_{IMU}^n \times \right) + \left( \mathbf{C}_b^n \left( \boldsymbol{\omega}_{nb}^b \times \right) \boldsymbol{l}_{wheel}^b \right) \times \right] \delta \boldsymbol{\psi} \\
&\quad - \mathbf{C}_n^v \mathbf{C}_b^n \left( \boldsymbol{l}_{wheel}^b \times \right) \delta \boldsymbol{\omega}_{ib}^b
\end{aligned} \quad (13)$$

where $\hat{\boldsymbol{v}}_{wheel}^v$ is the wheel velocity estimated by INS; $\boldsymbol{\omega}_{nb}^b$ is the angular rate vector of the $b$-frame with respect to the $n$-frame projected to the $b$-frame; $\hat{\boldsymbol{v}}_{IMU}^n$ is the INS-indicated IMU velocity; $\delta \boldsymbol{v}^n$ is the velocity error in the state vector; $\boldsymbol{l}_{wheel}^b$ indicates the lever arm vector between the Wheel-IMU and the $w$-frame projected in the $b$-frame; $\mathbf{C}_n^v$ can be obtained by Eq. (4) and Eq. (12); $\delta \boldsymbol{\psi}$ is the attitude error of the vehicle, which is only related to the heading error in the state vector. Thus, it can be written as $\delta \boldsymbol{\psi} = [0 \ 0 \ \delta \psi_b^n]^T$. Finally, the velocity error measurement equation in the $v$-frame can be written as

$$\begin{aligned}
\delta \boldsymbol{z}_v &= \hat{\boldsymbol{v}}_{wheel}^v - \tilde{\boldsymbol{v}}_{wheel}^v \\
&= \mathbf{C}_n^v \delta \boldsymbol{v}^n + \mathbf{C}_n^v \left[ \left( \mathbf{C}_b^n \left( \boldsymbol{\omega}_{nb}^b \times \right) \boldsymbol{l}_{wheel}^b \right) \times \right] \boldsymbol{\phi} \\
&\quad - \mathbf{C}_n^v \left[ \left( \boldsymbol{v}_{IMU}^n \times \right) + \left( \mathbf{C}_b^n \left( \boldsymbol{\omega}_{nb}^b \times \right) \boldsymbol{l}_{wheel}^b \right) \times \right] \delta \boldsymbol{\psi} \\
&\quad - \mathbf{C}_n^v \mathbf{C}_b^n \left( \boldsymbol{l}_{wheel}^b \times \right) \delta \boldsymbol{\omega}_{ib}^b
\end{aligned} \quad (14)$$

### B. Displacement Increment Measurement

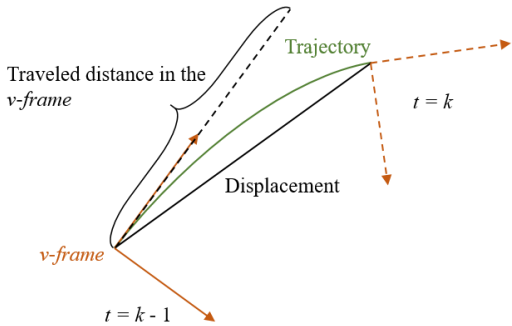

Fig. 3 Illustration of the displacement increment measurement (Top view).

The output of a standard odometer can be either the instantaneous vehicle velocity or the forward distance since last sampling time [20]. In this study, the gyroscope readings of the Wheel-IMU in the $x$-axis are leveraged to obtain wheel velocity at every IMU sampling moment. As opposed to the forward traveled distance in the $v$-frame indicated by the odometer [41], we exploit the displacement increment in the $n$-frame as observation, as shown in Fig. 3. The vehicle displacement increment in the $n$-frame is obtained by projecting the wheel velocity to the $n$-frame using the vehicle heading in the integral process. Because the odometer cannot perceive the change of the vehicle heading, the displacement increment model would more accurate and reliable than the forward distance, especially when vehicle is turning. In addition, the integral can mitigate the high-frequency noise of inertial sensor output to some extent.

According to Eq. (10) and Eq. (11), the velocity measurement in the $n$-frame ca be written as

$$\begin{aligned}
\tilde{\boldsymbol{v}}_{wheel}^n &= \hat{\mathbf{C}}_v^n \tilde{\boldsymbol{v}}_{wheel}^v - \boldsymbol{e}_v \\
&= \mathbf{C}_v^n \boldsymbol{v}_{wheel}^v + \left( \mathbf{C}_v^n \boldsymbol{v}_{wheel}^v \right) \times \delta \boldsymbol{\psi} - \boldsymbol{e}_v
\end{aligned} \quad (15)$$

The vehicle also has to be assumed to move on the horizontal surface here. Similar to Eq. (13), the INS-indicated velocity in the $n$-frame can be represented as

$$\begin{aligned}
\hat{\boldsymbol{v}}_{wheel}^n &= \hat{\boldsymbol{v}}_{IMU}^n + \hat{\mathbf{C}}_b^n \left( \hat{\boldsymbol{\omega}}_{nb}^b \times \right) \hat{\boldsymbol{l}}_{wheel}^b \\
&= \boldsymbol{v}_{wheel}^n + \delta \boldsymbol{v}^n + \left( \mathbf{C}_b^n \left( \hat{\boldsymbol{\omega}}_{nb}^b \times \right) \boldsymbol{l}_{wheel}^b \right) \times \boldsymbol{\phi} \\
&\quad - \mathbf{C}_b^n \left( \boldsymbol{l}_{wheel}^b \times \right) \delta \boldsymbol{\omega}_{ib}^b
\end{aligned} \quad (16)$$

Then the displacement measurement model is constructed by subtracting and integrating the two sides of Eq. (16) from Eq. (15) respectively. Assuming that the state errors keep constant within the integral time interval, we have

$$\begin{aligned}
\delta \boldsymbol{S}_v &= \int \hat{\boldsymbol{v}}_{wheel}^n - \int \tilde{\boldsymbol{v}}_{wheel}^n \\
&= \int \delta \boldsymbol{v}^n + \int \left( \mathbf{C}_b^n \left( \hat{\boldsymbol{\omega}}_{nb}^b \times \right) \boldsymbol{l}_{wheel}^b \right) \times \boldsymbol{\phi} \\
&\quad - \int \mathbf{C}_b^n \left( \boldsymbol{l}_{wheel}^b \times \right) \delta \boldsymbol{\omega}_{ib}^b - \int \left( \mathbf{C}_v^n \boldsymbol{v}_{wheel}^v \right) \times \delta \boldsymbol{\psi} + \boldsymbol{e}_S
\end{aligned} \quad (17)$$

where $\boldsymbol{e}_S$ is the measurement noise which is assumed as white Gaussian noise and

$$\begin{aligned}
\int \delta \boldsymbol{v}^n &= \frac{\delta \boldsymbol{v}_{k-1}^n + \delta \boldsymbol{v}_k^n}{2} \Delta t \\
&= \delta \boldsymbol{v}_k^n \Delta t - \frac{\delta \boldsymbol{v}_k^n - \delta \boldsymbol{v}_{k-1}^n}{2} \Delta t \\
&= \delta \boldsymbol{v}_k^n \Delta t - \frac{\Delta t}{2} \left( \int \delta \dot{\boldsymbol{v}}^n \right)
\end{aligned} \quad (18)$$

### C. Contact Point Zero-velocity Measurement

Fig. 4 shows the principle of constructing the contact point zero-velocity measurement. Unlike the velocity measurement model where the velocity of the Wheel-IMU is projected to the wheel center in the $v$-frame, in the contact point zero-velocity measurement model, the IMU velocity is projected to the contact point $p$ of the wheel in the $n$-frame. The measurement is built based on the fact that under general vehicle motion conditions (no slip and jumping), the velocity of the contact point on the wheel with respect to the ground is zero, namely, $\boldsymbol{v}_p^n = [0 \ 0 \ 0]^T$.

Let the wheel roll to the right with velocity $v_o$ and the angular rate of the wheel be $\omega$, then, the velocity of the contact point $p$ with respect to the wheel center $o$ is $v_p^o = \omega r$, pointing to left. Assuming that there is no slipping and jumping of the wheel, the magnitude of the velocity of the wheel center $v_o$ is equal to that of $v_p^o$ ($\omega r$), whereas the directions of them are opposite. Consequently, the velocity of the contact point $p$ with respect to the $n$-frame is zero. This scheme is similar to the foot-mounted IMU-based pedestrian navigation system [26, 42]. When a person is walking, his or her feet alternate between a stationary stance phase and a moving stride phase. Therefore, the foot-IMU can be used to detect the stance phase thereby the ZUPT can be performed to limit the error accumulation. While in our case, it is unnecessary to determine the stationary time because there is always a point on the wheel contacting with the ground.

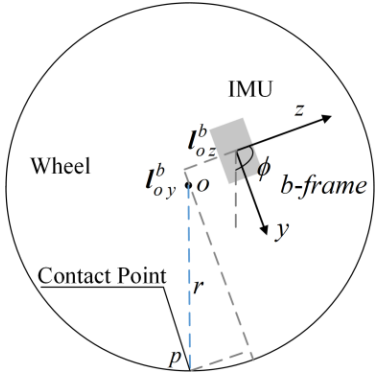

Fig. 4 Construction of the contact point zero-velocity measurement (Side view). $\phi$ is the roll angle of the IMU; $l_{oy}^b$ and $l_{oz}^b$ are the lever arm of the Wheel-IMU in the $y$- and $z$-axes, respectively; $r$ is the wheel radius; $o$ is the wheel center; $p$ is the contact point between the wheel and the ground.

The velocity of $p$ in the $n$-frame indicated by INS is
$$\hat{v}_p^n = \hat{v}_{IMU}^n + \hat{C}_b^n \left( \hat{\omega}_{ib}^b \times \right) \hat{l}_p^b \quad (19)$$
where $\hat{l}_p^b$ is the vector from the Wheel-IMU center $b$ to the contact point $p$ projected in the $b$-frame. Obviously, $\hat{l}_p^b$ changes periodically with the rotation of the wheel; thus, it should be calculated in real time, shown as follows
$$\hat{l}_p^b = \begin{bmatrix} l_{ox}^b \\ l_{oy}^b \\ l_{oz}^b \end{bmatrix} + r \begin{bmatrix} 0 \\ \sin\hat{\phi} \\ \cos\hat{\phi} \end{bmatrix} = \begin{bmatrix} l_{ox}^b \\ l_{oy}^b \\ l_{oz}^b \end{bmatrix} + r \begin{bmatrix} 0 \\ \sin(\phi+\delta\phi) \\ \cos(\phi+\delta\phi) \end{bmatrix}$$
$$\approx l_p^b + r \begin{bmatrix} 0 \\ \cos\phi \\ -\sin\phi \end{bmatrix} \delta\phi \quad (20)$$

where $\delta\phi$ is the roll angle error of the Wheel-IMU. As the same as the other two measurement models, the vehicle should also be assumed to move on the horizontal plane in the contact point zero-velocity measurement model, because the contact point determined by the Wheel-IMU is slightly different from the real contact point when the vehicle is moving uphill or downhill. Combining Eq. (19) and (20), the contact point zero-velocity measurement can be derived.

$$\delta z_{v,p} = \hat{v}_p^n - \tilde{v}_p^n$$
$$= v_{IMU}^n + \delta v^n - v_p^n$$
$$- (I - \phi\times) C_b^n \left( l_p^b \times + r \begin{bmatrix} 0 \\ \cos\phi \\ -\sin\phi \end{bmatrix} \delta\phi \times \right) \left( \omega_{ib}^b + \delta\omega_{ib}^b \right) + e_{vp} \quad (21)$$
$$= \delta v^n - \left[ \left( C_b^n (l_p^b \times) \omega_{ib}^b \right) \times \right] \phi$$
$$- C_b^n \left( \begin{bmatrix} 0 \\ r\cos\phi \\ -r\sin\phi \end{bmatrix} \times \right) \omega_{ib}^b \delta\phi - C_b^n (l_p^b \times) \delta\omega_{ib}^b + e_{vp}$$

where $e_{vp}$ is the measurement noise, modeled as white Gaussian noise.

Comparing with the other two measurements, the contact point zero-velocity measurement is more versatile and extensible, because all the ground vehicles, including wheeled robots, quadruped robots, and even pedestrians, have a point periodically contacting to the ground during their locomotion. Hence the contact point zero-velocity measurement can be straightforwardly utilized to correct the error drift of INS by mounting the IMU at an appropriate place of the vehicle to project its velocity to the contact point.

IV. EXPERIMENTAL RESULTS

This section provides and analyzes the experimental results to compare the performance of the proposed three different measurement models-based Wheel-INS. We evaluate the navigation performance of the three algorithms in both terms of positioning and heading through multiple sets of experiments with different vehicles and environments. Firstly, the experimental conditions and environments are described. Then, the performance comparison between the three measurement models is presented and analyzed.

A. Experimental Description

Field tests were conducted in three different places in Wuhan, China with two different ground vehicles. One was the Pioneer 3DX robot, a typical differential drive wheeled robot, and the other was a car. The Pioneer robot was used for two tests and the car for one. Fig. 5 shows the experimental platforms.

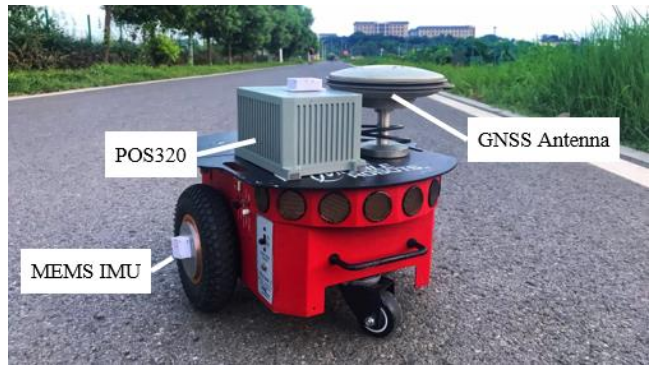

(a) Pioneer 3DX robot.

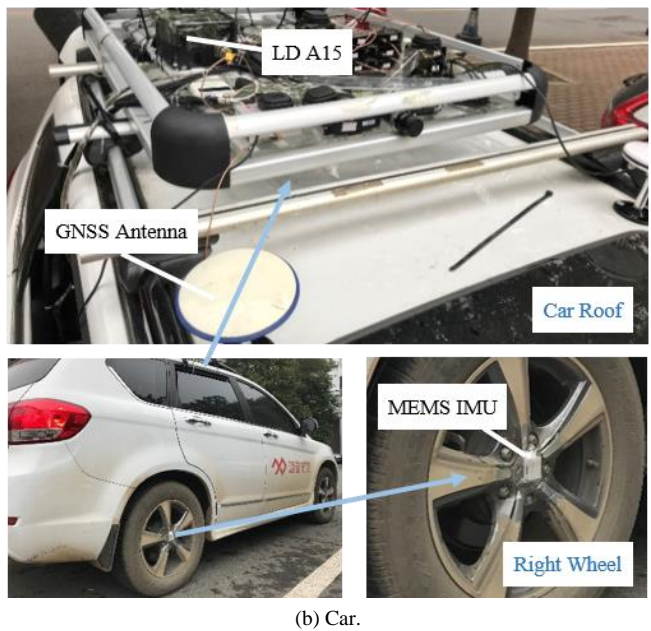

Fig. 5 Test platforms used in the real-world experiments.

TABLE I
TECHNICAL PARAMETERS OF THE IMUS USED IN THE EXPERIMENTS

| IMU | ICM20602 | POS320 | LD A15 |
|---|---|---|---|
| Gyro Bias ($\deg/h$) | 200 | 0.5 | 0.02 |
| Angle Random Walk ($\deg/\sqrt{h}$) | 0.24 | 0.05 | 0.003 |
| Accelerometer Bias ($m/s^2$) | 0.01 | 0.00025 | 0.00015 |
| Velocity Random Walk ($m/s/\sqrt{h}$) | 3 | 0.1 | 0.03 |

The MEMS IMU used in the experiments was a self-developed IMU module, containing four ICM20602 (TDK InvenSense) inertial sensor chips, a chargeable battery module, a microprocessor, a SD card for data collection, and a Bluetooth module for communication and data transmission. The IMU module can be connected with an android phone to record the raw data. We collected the outputs of two chips (logging at 200 Hz) in one trajectory as two sets of experimental data for post-processing. The MEMS IMU was carefully placed on the wheel to make them as close as possible to the wheel center. As shown in Fig. 5, the two vehicles were also equipped with two high-accuracy position and orientation systems to provide reference pose: POS320 (MAP Space Time Navigation Technology Co., Ltd., China) with a tactical-grade IMU for the robot experiments and LD A15 (Leador Spatial Information Technology Co., Ltd., China) with a navigation-grade IMU for the car experiments. Their main technique parameters are listed in TABLE I. The reference data were processed through a smoothed post-processed kinematic (PPK)/INS integration method. Technical references for generating the pose ground truth can be found in [43, 44]. The time synchronization between the MEMS IMU and the reference system was achieved via Bluetooth communication.

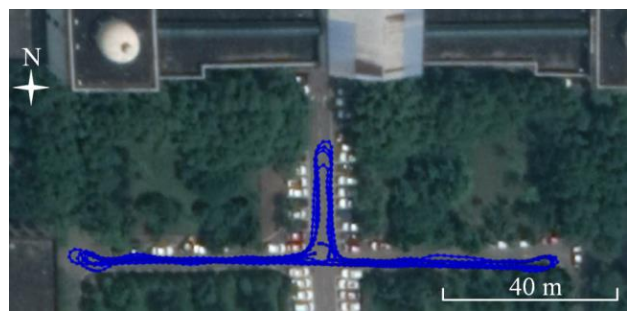

(a) Track I in the Information Department of Wuhan University.

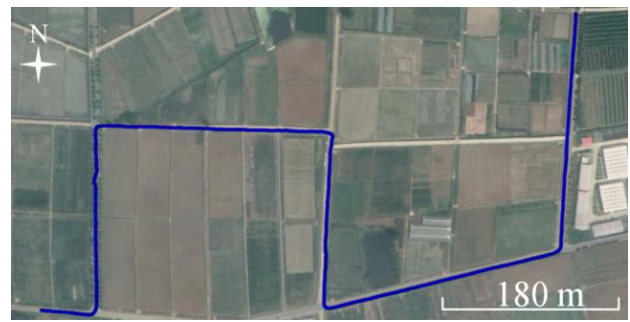

(b) Track II in the experimental farms in the Huazhong Agriculture University.

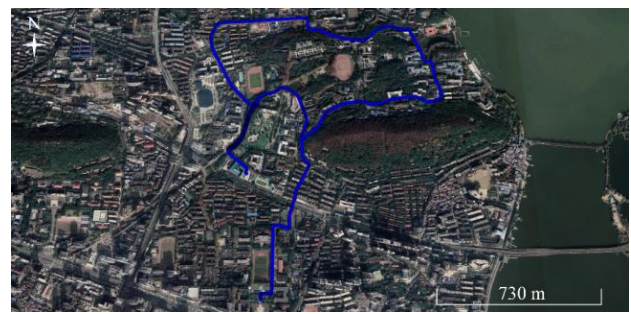

(c) Track III in the Wuhan University campus.

Fig. 6 Experimental trajectories.

TABLE II
VEHICLE MOTION INFORMATION IN THE EXPERIMENTS

| Test | Track | Vehicle | Average Speed (m/s) | Total Distance (m) |
|---|---|---|---|---|
| 1 | I | Pioneer 3DX | 1.39 | ≈1227 |
| 2 | | | | |
| 3 | II | | 1.25 | ≈1146 |
| 4 | | | | |
| 5 | III | Car | 4.70 | ≈12199 |
| 6 | | | | |

Fig. 6 shows the three test trajectories. Track I is a loopback trajectory in a small-scale environment in the Information Department of Wuhan University, on which the robot moved five times. Track II is a polyline trajectory with no return in the Huazhong Agriculture University. Track III is a large loop trajectory in the campus of Wuhan University, on which the robot moved approximately two times. The vehicle motion information of all the six tests is presented in TABLE II.

In our experiments, we used the approach proposed in [29] to calibrate and compensate the mounting angles before data processing. The lever arm was measured manually for three times to get the mean value. The initial heading, velocity, and position of Wheel-INS were given by the reference system directly. We chose this simple method for the initial alignment of INS because we mainly focused on the DR performance of Wheel-INS. However, other alignment methods should be investigated for practical applications. The static IMU data before the vehicle started moving were used to estimate the initial roll and pitch, as well as the initial gyroscope bias of the Wheel-IMU. The initial values of other inertial sensor errors were set as zero. The update frequency was set as 2 Hz in all the three measurement models-based Wheel-INS.

In our previous research on Wheel-INS [24], we have illustrated the advantages of Wheel-INS in terms of DR performance and resilience to the gyroscope bias through extensive field experiments. Therefore, in this paper, the experimental analysis mainly focuses on the comparison of the three measurement models in Wheel-INS.

### B. Performance Comparison of the Three Measurements

The positioning error in the horizontal plane and the heading error of the three measurement models in Test 1 and Test 5 are presented in Fig. 7.

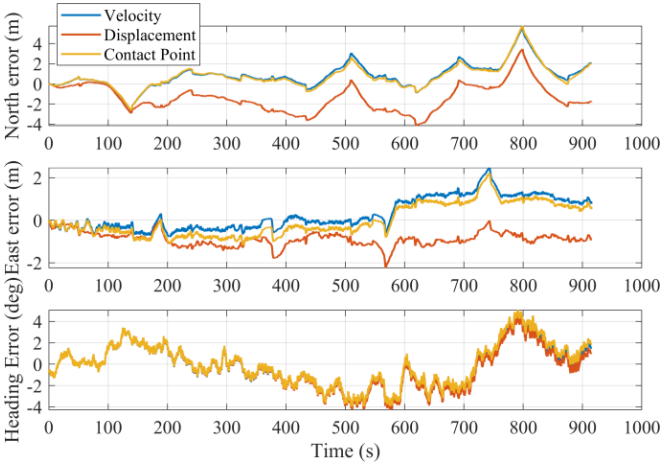

(a) The position and heading errors of the three measurement models in Test 1.

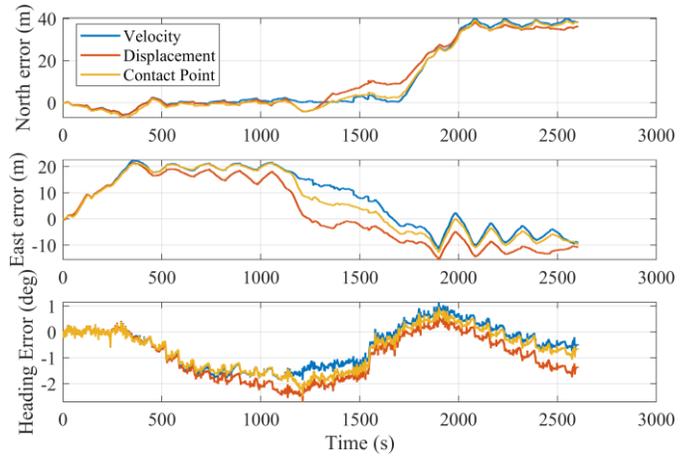

(b) The position and heading errors of the three different models in Test 5.

Fig. 7 The positioning errors in the north and east direction, and the heading error of the three measurement models-based Wheel-INS in Test 1 and Test 5. "Velocity", "Displacement", and "Contact Point" indicate the velocity measurement, the displacement increment measurement, and the contact point zero-velocity measurement, respectively.

It's obvious in Fig. 7 that there is no significant difference between the navigation errors of the three measurement models-based Wheel-INS on the whole. They show similar positioning and heading error drift in views of both the whole and local trajectory. However, it can be observed in Fig. 7 (a) that the displacement increment measurement-based Wheel-INS shows different drift trend in about 140s in Test 1. This can be considered as a stochastic phenomenon owing to the random error since we have processed the data from other IMU chips inside the same IMU module in Test 1 and this is not always the case.

It is common to calculate the maximum position drift of the entire trajectory or the misclosure error to evaluate the positioning performance of a DR system in the community. However, this metric is not strict because the loop of the trajectory will suppress error accumulation to some extent, especially for INS in which the positioning error always drifts in one direction. For example, it can be observed from Fig. 7 (a) that when the robot turns around, the positioning error starts to drift along the opposite direction. Therefore, we use the mean drift rate as the evaluation ariterion here. Firstly, we accumulated the traveled distance of the vehicle by a certain increment ($l$) and calculated the horizontal position error drift rate (equaling to the maximum horizontal positioning error in current traveled distance divided by the traveled distance) within each distance ($l, 2l, 3l,...$). Then, the mean value (MEAN) and standard deviation (STD, $1\sigma$) were computed as the final indicator of positioning performance. This approach is similar to the odometry evaluation metric proposed in the KITTI dataset [45], but we segmented the trajectory only from the starting point. With regard to the heading error, the maximum (MAX) and root mean square error (RMSE) were calculated. In this work, we chose $l$ as 100 m. Fig. 8 and Fig. 9 show the position drift rate in the horizontal plane of the three systems (which is a function of the traveled distance) in Test 1 and Test 5, respectively.

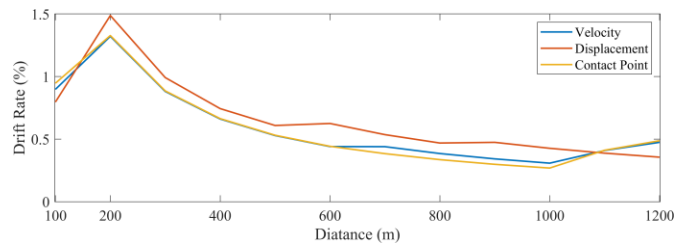

Fig. 8 The horizontal positioning drift rate of the three systems in Test 1.

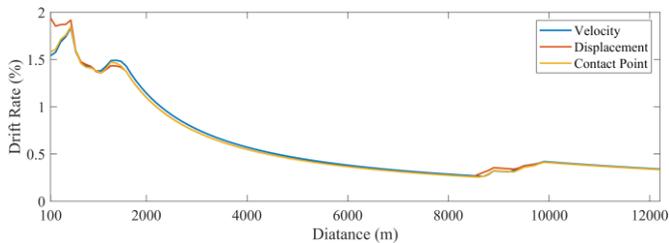

Fig. 9 The horizontal positioning drift rate of the three systems in Test 5.

It can be observed that the drift rates of the three algorithms are very close. In addition, the position drift rates of the three measurements all present a downward trend with the increase of the distance. This is due to that the loop closure in Track I and Track III suppressed the position drift.

TABLE III lists the error statistics of the three systems in all the six experiments. Fig. 10 draws the error statistics of the three measurements –based Wheel-INS in all the six testes.

TABLE III
DR PERFORMANCE COMPARISON OF THE THREE MEASUREMENTS

| Test No. | Measurement | Position Drift Rate (%) | | Heading Error (°) | |
|---|---|---|---|---|---|
| | | MEAN | STD | MAX | RMSE |
| 1 | Velocity | 0.59 | 0.30 | 4.79 | 1.93 |
| | Displacement | 0.66 | 0.32 | 4.50 | **1.91** |
| | Contact Point | **0.58** | 0.32 | 5.06 | 1.93 |
| 2 | Velocity | 1.43 | 0.54 | 7.93 | 3.88 |
| | Displacement | 1.66 | 0.98 | 7.63 | 3.26 |
| | Contact Point | **1.34** | 0.58 | 7.03 | **2.70** |
| 3 | Velocity | 1.17 | 0.27 | 4.56 | 2.16 |
| | Displacement | **0.96** | 0.24 | 4.47 | **2.15** |
| | Contact Point | 1.32 | 0.33 | 4.50 | 2.16 |
| 4 | Velocity | 1.78 | 0.26 | 10.88 | 4.44 |
| | Displacement | 1.87 | 0.35 | 9.34 | **4.18** |
| | Contact Point | **1.76** | 0.43 | 10.83 | 4.94 |
| 5 | Velocity | 0.62 | 0.42 | 1.91 | **0.96** |
| | Displacement | 0.61 | 0.44 | 2.70 | 1.28 |
| | Contact Point | **0.60** | 0.42 | 2.48 | 1.03 |
| 6 | Velocity | 0.83 | 0.43 | 4.97 | 1.60 |
| | Displacement | **0.61** | 0.50 | 2.55 | 1.22 |
| | Contact Point | 0.66 | 0.47 | 3.53 | **1.00** |

From TABLE III, we can learn that in all the six experiments, the horizontal position drift rates of all the three measurements-based Wheel-INS are all less than 2%. And the RMSE of the heading error are all less than 5°.

It is evident in Fig. 10 that the three measurements show an equivalent navigation performance. It is hard to determine which measurement model under what conditions can achieve a better performance the other two. For instance, the displacement increment measurement slightly outperforms the other two measurements in Test 3, while the contact point zero-velocity measurement generates the best position estimation in Test 1 and Test 2. As for the heading accuracy, the three systems also show a same level of accuracy in each experiment.

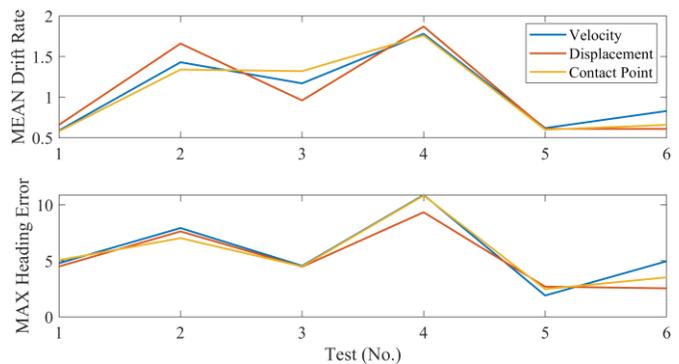

Fig. 10 MEAN position drift rate and MAX heading error of the three methods in all the six tests.

Basically, the three measurements leverage the same information (wheel velocity) to construct the observation models. In addition, all the three kinds of measurements rely on the NHC and the horizontal motion assumption of the vehicle. The velocity measurement utilizes the velocity of the wheel center along with the NHC to fuse with INS directly. The displacement increment measurement integrates the wheel velocity in a short time interval to obtain the incremental displacement in the *n*-frame to suppress the error drift of INS. And the contact point zero-velocity measurement projects the velocity to the contact point between the wheel and the ground, so as to construct the constraint.

However, different measurement model would cause different error. For example, the displacement increment model is affected by the heading error because vehicle heading is required to project the forward distance of the vehicle to the *n*-frame at every IMU data epoch, while the rolling angle error of the Wheel-IMU would be introduced in the velocity projection process in the contact point zero-velocity measurement. It is evident that the sensor errors (e.g., random noise) of low-cost MEMS IMUs are much more significant than the modeling error. Moreover, the rotation of the wheel would eliminate a large part of the heading gyroscope bias error, which is one of the main error sources of INS. As a result, these observation information would contribute limitedly to improve the heading accuracy. In conclusion, the DR performance of the three measurements-based Wheel-INS using a MEMS IMU should not be dramatically different.

## V. DISCUSSION

From the derivations of the three measurement models in Section III, it can be learned that all the three types of observations leverage the same vehicle motion information to construct the measurement models: the vehicle forward velocity and NHC. Although each algorithms exhibits its own pros and cons, the navigation performance on the whole is at the same level. However, it is worth mentioning that the residual lever arm error has less impact on the displacement increment measurement because it integrates the velocity within a certain time interval as observation rather than the instantaneous velocity.

To investigate the influence of the residual lever arm error on the three measurement models, we manually added a bias in the measured lever arm and then compared the positioning errors of the three systems in Test 1. Because the misalignment error in the wheel plane (namely, the lever arm in $y$- and $z$-axes of the $b$-frame) are more important, we only added errors in these two direction, which were both set as 0.2 cm. Fig. 11 shows the corresponding positioning and heading errors of the three systems in Test 1.

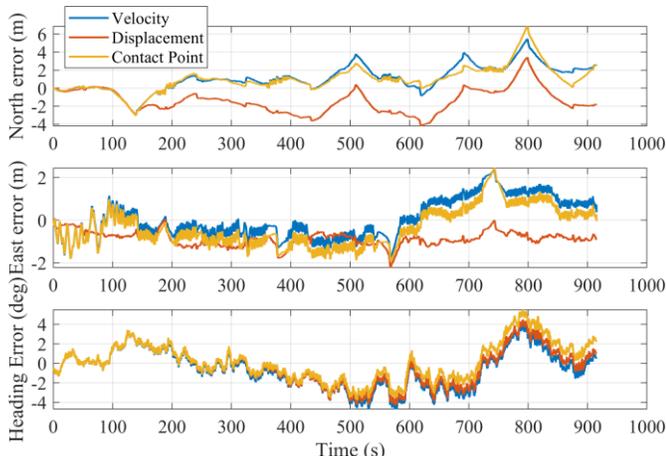

Fig. 11 The positioning error in the north and east direction, and the heading error of the three systems in Test 1 (with additional lever arm error).

Comparing with Fig. 7(a), it can be observed that the positioning errors of the velocity measurement and the contact point zero-velocity measurement-based Wheel-INS have increased. Besides, there is an obvious vibration in the positioning error of these two systems. In the velocity measurement model and the contact point zero-velocity measurement model, the lever arm is essential to project the velocity of the Wheel-IMU to the reference point (wheel center and contact point). Note that the positioning error of Wheel-INS caused by the residual lever arm error is mainly embodied in the forward direction of the vehicle because with the rotation of the wheel, the velocity projection error changes its direction around the rotation axis periodically. As a result, the positioning errors in the vehicle direction vibrate significantly in these two systems (the velocity measurement and contact point zero-velocity measurement-based Wheel-INS). However, with the integration of the velocity in the displacement increment measurement, the periodical velocity projection error caused by the residual lever arm error would be cancelled to some extent; thus it would not lead to evident deterioration in the final positioning results. In conclusion, the displacement increment measurement exhibits a desirable immunity to the lever arm error.

## VI. CONCLUSION

In this article, a wheel-mounted MEMS IMU-based DR system is studied. Particularly, three types of measurement models are exploited based on the Wheel-IMU, including the velocity measurement, displacement increment measurement, and contact point zero-velocity measurement. Basically, the observation information utilized in all the three measurements is the same: wheel velocity. Although different errors are introduced when different measurement models are constructed, they are trivial compared to the sensor errors of the MEMS IMU. Furthermore, a large part of the heading gyroscope bias error, which is one of the main error sources of INS, can be canceled with the rotation of the wheel. Therefore, the final navigation results of Wheel-INS based on the three measurement models should be at the same level.

Field tests with different vehicle platforms in different environments illustrate the feasibility and equivalence of the proposed three measurement models. The maximum horizontal position drifts are all less than 2% of the total travelled distance. Nonetheless, there are some specific characteristics of these measurements. Firstly, the displacement increment measurement shows considerable insensitivity to the lever arm error comparing with the other two measurements. Secondly, the velocity measurement is more straightforward and concise to be implemented. Finally, the contact point zero-velocity measurement exhibits better versatility for different kinds of ground vehicles. We have made the example data and code available to the community (https://github.com/i2Nav-WHU/Wheel-INS).

Although Wheel-INS can provide considerable DR results, the positioning errors will inevitably accumulate because of the lack of external correction information. For the future research, integrating other exteroceptive sensors (e.g., camera and LiDAR) to enable the loop closure would be a promising approach to eliminate the long-term error accumulation.

### ACKNOWLEDGEMENTS

The authors would like to thank Dr. Jussi Collin for inspiring us to investigate the contact point zero-velocity measurement model in our private communication.